%% file: main.tex
\documentclass[10pt,twocolumn,letterpaper]{article}

\usepackage[pagenumbers]{cvpr} 

\input{preamble}

\definecolor{cvprblue}{rgb}{0.21,0.49,0.74}
\usepackage[pagebackref,breaklinks,colorlinks,citecolor=cvprblue]{hyperref}

\def\paperID{9342} 
\def\confName{CVPR}
\def\confYear{2024}

\title{HPNet: Dynamic Trajectory Forecasting with Historical Prediction Attention}

\author{Xiaolong Tang\textsuperscript{\rm 1,2}
\quad
Meina Kan\textsuperscript{\rm 1,2}
\quad
Shiguang Shan\textsuperscript{\rm 1,2}
\quad
Zhilong Ji\textsuperscript{\rm 3}
\quad
Jinfeng Bai\textsuperscript{\rm 3}
\quad
Xilin CHEN\textsuperscript{\rm 1,2}\\
\textsuperscript{\rm 1}Institute of Computing Technology, Chinese Academy of Sciences\\
\textsuperscript{\rm 2}University of Chinese Academy of Sciences
\quad\quad \textsuperscript{\rm 3} Tomorrow Advancing Life\\
{\tt\small tangxiaolong22s@ict.ac.cn}
}

\begin{document}
\maketitle
\input{sec/0_abstract}    
\input{sec/1_introduction}
\input{sec/2_related_work}
\input{sec/3_method}
\input{sec/4_experiment}
\input{sec/5_conclusion}
\clearpage
{
    \small
    \bibliographystyle{ieeenat_fullname}
    \bibliography{main}
}

\setcounter{section}{0}
\input{sec/X_suppl}
\end{document}

%% file: preamble.tex
\usepackage{multirow} 
\usepackage[dvipsnames]{xcolor}


%% file: sec/0_abstract.tex
\begin{abstract}
Predicting the trajectories of road agents is essential for autonomous driving systems. The recent mainstream methods follow a static paradigm, which predicts the future trajectory by using a fixed duration of historical frames. These methods make the predictions independently even at adjacent time steps, which leads to potential instability and temporal inconsistency. As successive time steps have largely overlapping historical frames, their forecasting should have intrinsic correlation, such as overlapping predicted trajectories should be consistent, or be different but share the same motion goal depending on the road situation. Motivated by this, in this work, we introduce HPNet, a novel dynamic trajectory forecasting method. Aiming for stable and accurate trajectory forecasting, our method leverages not only historical frames including maps and agent states, but also historical predictions. Specifically, we newly design a Historical Prediction Attention module to automatically encode the dynamic relationship between successive predictions. Besides, it also extends the attention range beyond the currently visible window benefitting from the use of historical predictions. The proposed Historical Prediction Attention together with the Agent Attention and Mode Attention is further formulated as the Triple Factorized Attention module, serving as the core design of HPNet. 
Experiments on the Argoverse and INTERACTION datasets show that HPNet achieves state-of-the-art performance, and generates accurate and stable future trajectories.
Our code are available at \url{https://github.com/XiaolongTang23/HPNet}.

\end{abstract}

%% file: sec/1_introduction.tex
\section{Introduction}
\label{sec:introduction}

\begin{figure}[t]
  \centering
  \includegraphics[width=0.9\linewidth]{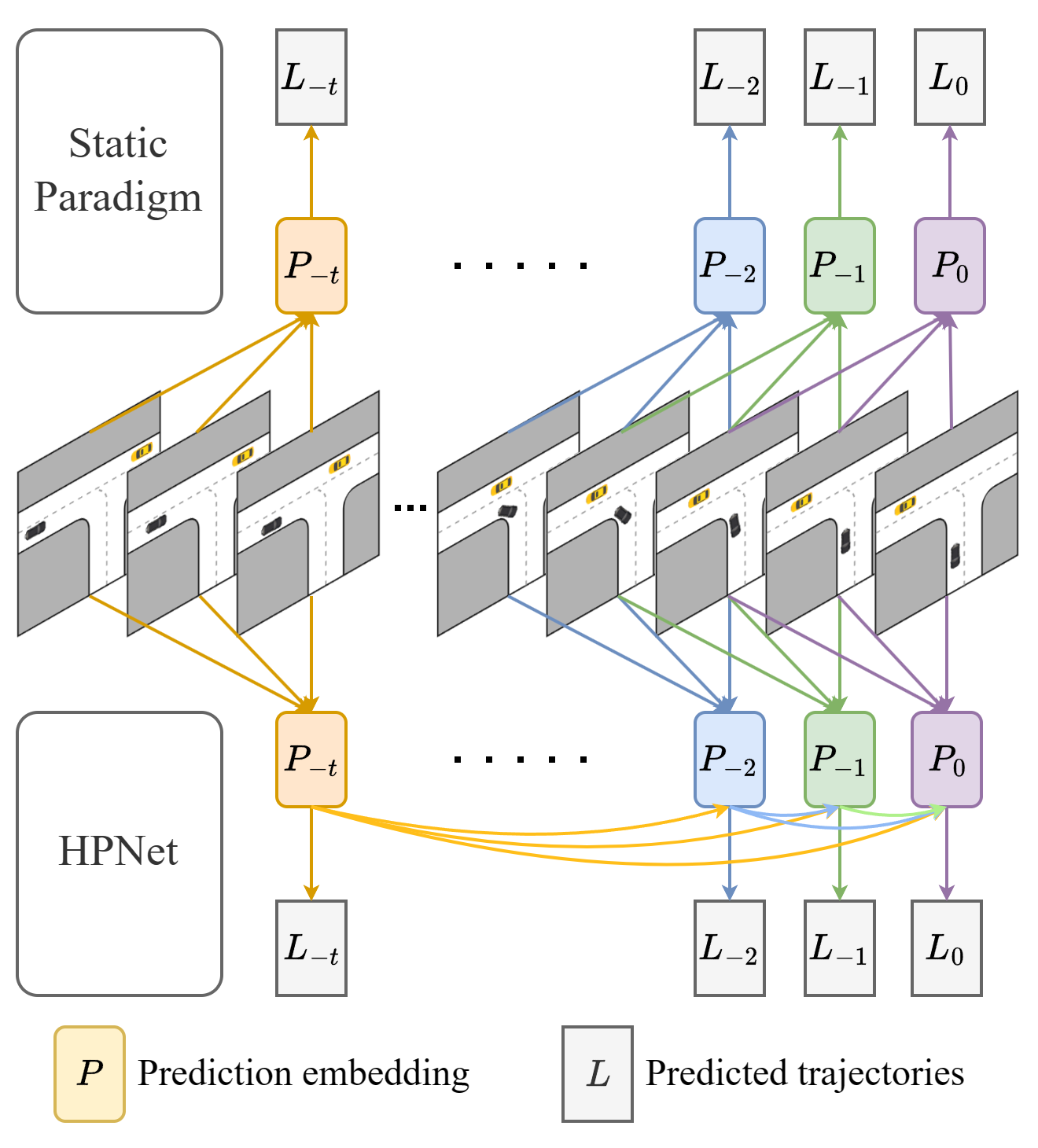}
   \caption{The difference between previous methods and ours. Previous methods (upper) treat trajectory prediction as a static task, predicting future trajectories based on a fixed-length sequence of historical frames. They independently forecast trajectories even at adjacent timesteps, despite the considerable overlap in input data. In contrast, HPNet (lower) views trajectory prediction as a dynamic task. It not only leverages historical frames but also historical prediction embeddings to forecast trajectories.}
   \label{fig:contrast}
\end{figure}

Accurate and reliable trajectory prediction of road agents such as cars and pedestrians is critical to the decision-making and safety of autonomous driving systems. However, trajectory prediction is extremely challenging. On the one hand, an agent's motion is influenced not only by road geometry and rules but also by surrounding agents. On the other hand, the agent's intentions are unknown, leading to high levels of future uncertainty.

Recently, researches such as Macformer \cite{feng2023macformer}, HiVT \cite{zhou2022hivt},  and Multipath++ \cite{varadarajan2022multipath++} have achieved notable results by employing intricately designed network architectures to seamlessly fuse heterogeneous information including agent history, agent-agent interactions, and agent-map interactions. Wayformer \cite{nayakanti2023wayformer} further explores the unified architecture for fusing the heterogeneous information. In addition, to account for future uncertainties, recent work \cite{chai2020multipath, phan2020covernet, zhao2021tnt, gu2021densetnt, wang2022ltp, zhang2021map, ngiam2021scene, liu2021multimodal, varadarajan2022multipath++, nayakanti2023wayformer, zhou2023query, wang2023prophnet, kim2021lapred} have shifted towards generating multi-modal future trajectories rather than single trajectory, since road agents may make varying decisions even in identical scenarios.       Anchor-based approaches \cite{chai2020multipath, zhao2021tnt, zhang2021map, gu2021densetnt, wang2022ltp} utilize multiple candidate goals or predefined paths as anchors to indicate various potential future, thereby facilitating the generation of multi-modal trajectories.      More recently, models \cite{liu2021multimodal, varadarajan2022multipath++, nayakanti2023wayformer, zhou2023query, wang2023prophnet} adopt learnable queries to generate multi-modal predictions, achieving promising results.

While the existing methods have achieved great advancements in prediction accuracy, they mostly treat trajectory prediction as a static task by using a fixed number of historical frames to predict future trajectories. As shown in \cref{fig:contrast}, successive predictions are inherently independent, although their input overlaps considerably. \textit{This static paradigm of trajectory prediction may lead to instability and temporal inconsistency in successive predictions, which is not conducive to the autonomous driving system making safe and reliable decisions.} Aiming for more stable prediction, DCMS \cite{ye2022dcms} proposes to model trajectory prediction as a dynamic problem. It explicitly considers the correlation between successive predictions and imposes a temporal consistency constraint that requires the overlapping parts of predicted trajectories at adjacent time steps to be identical. Moreover, QCNet \cite{zhou2023query} introduces a query-centric encoding paradigm to encode location-dependent features and location-independent features separately, thereby circumventing redundant encoding in successive predictions. These preliminary studies show the effectiveness and rationality of modeling trajectory prediction as a dynamic task. 

These works \cite{ye2022dcms, zhou2023query} inspire us to think that the intrinsic relationship between successive predictions should be more general, not just consistent. For example, the overlapping portions may remain consistent as that in DCMS \cite{ye2022dcms} or change slightly. Even more, when an agent is navigating through a congested multi-way intersection, the successive predictions may be quite different but still share the same motion goal as shown in \cref{fig:w/o HPA and w/ HPA} (b). So, in this work, we present a novel dynamic trajectory prediction method, HPNet. It models the dynamic relationship between successive predictions as the process of History Prediction Attention. Specifically, HPNet consists of three components: Spatio-Temporal Context Encoding, Triple Factorized Attention, and Multimodal Output. Initially, the mode queries aggregate spatio-temporal context to form preliminary prediction embeddings. Subsequently, Triple Factorized Attention, comprising Agent Attention, Historical Prediction Attention, and Mode Attention, models the interactions between agents, predictions, and modes, respectively, to obtain more informative prediction embeddings. Finally, the embeddings are decoded as multimodal future trajectories in the last module.

Our method has two clear advantages: First, our model establishes a general relationship between successive predictions, using the historical predictions as references to improve stability and increase accuracy. Second, in online inference, existing static attention-based methods are constrained within the fixed visible historical range, due to limited dataset size or computational resources. Instead, our approach can achieve a larger visible range (i.e. longer attention) without increasing computational overhead, which is beneficial for better accuracy in practical applications.

%% file: sec/2_related_work.tex
\section{Related Work}
\label{sec:related work}

\begin{figure*}
    \centering
    \includegraphics[width=1.0\linewidth]{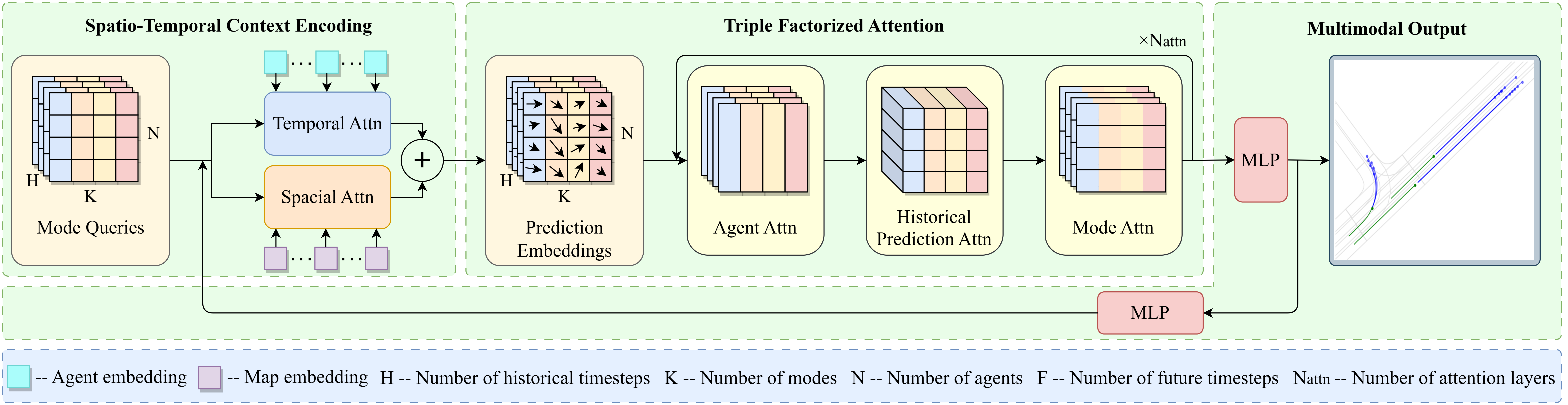}
    \caption{An overview of HPNet. The proposed HPNet encompasses three components: Spatio-Temporal Context Encoding, Triple Factorized Attention, and Multimodal Output.      Firstly, it combines agent and lane features with mode queries to create initial prediction embeddings.      Subsequently, Triple Factorized Attention — comprising Agent Attention, our proposed Historical Prediction Attention, and Mode Attention — refine these prediction embeddings.  Finally, the prediction embeddings are decoded by an MLP to obtain the predicted trajectories. The predicted trajectories are fed into this pipeline again to enhance the precision of predictions.}
    \label{fig:HPNet}
\end{figure*}

\textbf{Attention Mechanism.}
Transformer \cite{vaswani2017attention} has achieved notable success in fields such as natural language processing and computer vision. This is largely attributed to the attention mechanism, which considers the whole context and focuses on the important parts of the input data. Recently, many methods \cite{liu2021multimodal, gilles2022thomas, zhou2022hivt, wang2023prophnet, zhou2023query, ngiam2021scene, nayakanti2023wayformer, cheng2023forecast, yuan2021agentformer, zhang2024real, choi2023r, girgis2021latent, jia2023hdgt, zhu2023biff, seff2023motionlm} employ attention to process agent history sequence or model agent-agent and agent-lane interactions, and have achieved great success in trajectory prediction. The input for the trajectory prediction task usually encompasses data across temporal and spatial dimensions. Instead of flattening the input together into one joint self-attention \cite{wang2023prophnet, cheng2023forecast, yuan2021agentformer}, applying attention to each axis
\cite{liu2021multimodal, zhou2022hivt, girgis2021latent, zhou2023query, ngiam2021scene, nayakanti2023wayformer} may be more suitable for trajectory prediction task, which leads to better semantic consistency as well as lower computational complexity. In our work, the key module Triple Factorized Attention also follows the latter attention approach, consisting of Historical Prediction Attention, Agent Attention, and Mode Attention to model the interaction of agents in three different dimensions.

\textbf{Multimodal Output.}
To model future uncertainty distributions, probabilistic methods \cite{deo2022multimodal, yuan2021agentformer, gupta2018social, sadeghian2019sophie, casas2020implicit, salzmann2020trajectron++} firstly propose to use generative models (\textit{e.g.}, GAN, VAE) to obtain multimodal outputs through multiple samplings. However, in these methods, the number of sampling iterations required to produce reliable results is indeterminate, making them unreliable for the later decision process of autonomous driving. Afterwards, deterministic methods \cite{zhao2021tnt, gu2021densetnt, chai2020multipath, varadarajan2022multipath++, wang2022ltp, liu2021multimodal, zhou2022hivt, ngiam2021scene, wang2023prophnet, zhou2023query, ye2022dcms, nayakanti2023wayformer, liang2020learning, ye2021tpcn, shi2022motion, li2023fsr, kim2021lapred, wang2023ganet, phan2020covernet} propose to eliminate the need of multiple sampling and directly output multi-modal future trajectories in a single shot, yielding more accurate trajectory prediction. Among them, Anchor-based approaches usually employ two types of anchors, candidate targets \cite{zhao2021tnt, gu2021densetnt, wang2022ltp, kim2021lapred} and predefined paths \cite{chai2020multipath, zhang2021map, phan2020covernet}. These anchors indicate a variety of potential future trajectories, thus effectively improving the prediction accuracy of multimodal trajectories. However, the performance of these methods heavily depends on the quality of the anchors, and sometimes bad anchors can lead to irreparably bad results. So, inspired by DETR \cite{carion2020end}, several methods propose to use learnable queries \cite{liu2021multimodal, varadarajan2022multipath++, nayakanti2023wayformer, zhou2023query, wang2023prophnet, shi2022motion} rather than anchors to facilitate multimodal output. In these methods, each mode query can adaptively generate a potential trajectory for every sample based on its context, which is more flexible than pre-defined fixed anchors, leading to promising performance in trajectory prediction. Lately, there appear some approaches \cite{zhu2023biff, shi2022motion, choi2023r, wang2023prophnet, zhou2023query} to combine query-based and anchor-based methods. For example, ProphNet \cite{wang2023prophnet} generates specific anchors based on the sample and feeds the anchor information into learnable mode queries to help produce multimodal trajectories. QCNet \cite{zhou2023query} adaptively generates trajectory proposals through anchor-free mode queries and then refines these proposals based on the context using anchor-based mode queries.

%% file: sec/3_method.tex
\section{Method}
\label{sec:method}
Trajectory forecasting aims to predict the future trajectories of any agent given its historical status. Specifically, given a fixed length sequence of history status frames 
 \(\{f_{-T+1}, f_{-T+2}, ..., f_0\}\),  the goal is to predict \( K \) different modal trajectories for \(N\) agents as below:
\begin{equation}
    L_0 = \{L_{0,n,k}\}_{n\in [1,N], k\in[1,K]},     
    \label{eq:prediction goal}
\end{equation}
where \( f_t=\{a_t^{1\sim N}, \mathcal{M}\} \), \( a_t^{1\sim N} \) represents the features of all agents in the scene at time \( t \), and \( \mathcal{M} \) denotes the high-definition (HD) map including \( M \) lane segments. And each trajectory contains future locations for the next $F$ time steps:
\begin{equation}   
    L_{0,n,k}=\{l_{1,n,k}, l_{2,n,k}, ..., l_{F,n,k}\}, 
    \label{eq:detailed prediction goal}
\end{equation}
where $l_{i,n,k}\in \mathbb{R}^2$ represents the predicted position at time step \(i\) of mode \(k\) for agent \(n\). Simultaneously, a probability score is usually obtained for each predicted trajectory to indicate its likelihood of being the path that the agent will actually follow. 

An overview of our proposed HPNet is illustrated in \cref{fig:HPNet}. As shown, our model consists of three parts: Spatio-Temporal Context Encoding, Triple Factorized Attention, and Multimodal Output. Firstly, the spatio-temporal features of agents and lanes are aggregated with learnable mode queries to generate prediction embeddings that can preliminarily predict future trajectories. Then, Triple Factorized Attention including Agent Attention, Historical Prediction Attention, and Mode Attention are conducted to refine the prediction embeddings. Among them, Agent Attention models interactions between agents, Mode Attention models interactions across different modes (\textit{i.e.}, different predicted paths), and Historical Prediction Attention is a novel module we propose to dynamically model the intrinsic correlation between current and historical predictions. Finally, The prediction embeddings are decoded by a MLP to obtain the predicted trajectories, which are fed into the whole pipeline again to enhance the precision of forecasts.

\subsection{Spatio-Temporal Context Encoding}
\label{subsec:Spatio-Temporal Context Encoding}
HPNet is based on Graph Neural Networks (GNNs) and adopts relative spatio-temporal position encoding \cite{zhou2023query, jia2023hdgt, zhu2023biff, zhang2024real}. It encodes the location-independent features of agents and maps into node embeddings while encoding the relative spatio-temporal positions into edge embeddings.

\textbf{Encoding Agent Features.}
The agent features include spatial position, motion state, and semantic attributes of each agent at each time step. Each agent at each time step is adopted as a node in the graph, and its features represented as \( a_t^n=\{p^{t,n}_x,p^{t,n}_y,\theta^{t,n},v^{t,n}_x,v^{t,n}_y,c^{t,n}_a\} \), where \( (p^{t,n}_x,p^{t,n}_y) \) is the location, \( \theta^{t,n} \) is the orientation, \( (v^{t,n}_x,v^{t,n}_y) \) is the velocity, and \( c^{t,n}_a \) is the attribute. For each agent at each time step, we take its location as the origin of a local polar coordinate system and its orientation as the positive direction. In this reference frame, the velocity \( (v^{t,n}_x,v^{t,n}_y) \) is represented as \( (v^{t,n}, \varphi^{t,n}) \) where \( v^{t,n} \) is the velocity magnitude and \( \varphi^{t,n} \) is the direction of velocity. A two-layer MLP is adopted to encode the location-independent features as agent embeddings \( E^{t,n}_a=\mathrm{MLP}(v^{t,n}, \varphi^{t,n}, c^{t,n}_a) \), where \( E^{t,n}_a \in \mathbb{R}^{D} \), \(D\) is the encoding dimension.

\textbf{Encoding Map Features.}
The map features include spatial position, length, and semantic attributes of each lane segment. The lane segments are adopted as nodes in the graph, each of which comprises a set of centerlines along with some attributes.  The location and orientation of the midpoint of centerlines are used to represent the position and orientation of each lane segment.  The lengths of the lane segments \( l_ m\) and their attributes \( c_m \) serve as node features, which are encoded through a two-layer MLP as map embeddings \( E_m=\mathrm{MLP}(l_m,  c_m) \), where \( E_m \in \mathbb{R}^{M \times D} \). Similar to LaneGCN \cite{liang2020learning}, to capture the topological structure of the map, the lane nodes are connected based on adjacent, predecessor, and successor relationships. Interaction between lane nodes is then accomplished via self-attention.

\textbf{Encoding Relative Spatio-Temporal Position.}
The relative spatio-temporal positions between nodes are used as the features for the edges. All nodes in the graph are encoded using features in their local polar coordinates, so the edges represent the transformation relationships between different local polar coordinates. The edge features can be denoted as \( \{d_e,\phi_e,\psi_e,\delta_e\} \), where \( d_e \) represents the distance from the source node to the target node, \( \phi_e \) indicates the orientation of the edge in the target node's reference frame, \( \psi_e \) denotes the relative orientation between the source and target nodes, and \( \delta_e \) signifies the time difference between them. Similarly, a two-layer MLP is adopted to encode these features into edge embeddings \( E_e=\mathrm{MLP}(d_e,\phi_e,\psi_e,\delta_e) \), where \( E_e \in \mathbb{R}^{Y\times D} \), \(Y\) is the number of edges.

\textbf{Spatio-Temporal Attention.}
Spatio-temporal attention comprises two parallel cross-attention modules. Temporal Attention aggregates the historical embeddings of agents, while Spatial Attention models agent-lane interaction. We assign the same learnable mode queries to each agent at each time step, denoted as \( \{q_{t,n,k}\}_{t\in[1-T,0],n\in[1, N],k\in[1, K]} \). Each mode query, as a node in the graph, the spatio-temporal position of which is identical to the corresponding agent. For each mode query node, Spatial Attention with lane nodes is performed within a certain spatial radius \(R_1\) and Temporal Attention with agent nodes within a specified time span \(I_1\), respectively. Edges participate in this process via concatenating with the source node \cite{zhou2022hivt, jia2023hdgt, zhou2023query, zhu2023biff, zhang2024real}:
\begin{equation}
  q_{t,n,k}^S = \mathrm{MHA}(q_{t,n,k}, [E_m, E_e], [E_m, E_e]),
  \label{eq:spatial attention}
\end{equation}
\begin{equation}
  q_{t,n,k}^T = \mathrm{MHA}(q_{t,n,k}, [E_a^{t-I_1\sim t, n}, E_e], [E_a^{t-I_1\sim t, n}, E_e]),
  \label{eq:temporal attention}
\end{equation}
where \(\mathrm{MHA}(a, b, c)\) denotes the multi-head attention with \(a\) as query, \(b\) as key, and \(c\) as value.The results of the two cross-attention modules are then summed to generate the prediction embeddings:
\begin{equation}
  P_{t,n,k} = q_{t,n,k}^T + q_{t,n,k}^S.
  \label{eq:prediction embedding}
\end{equation}

Subsequently, the prediction embeddings generated from \cref{eq:prediction embedding} are passed through Triple Factorized Attention. Triple Factorized Attention comprises Agent Attention, Historical Prediction Attention, and Mode Attention, allowing each prediction embedding to directly or indirectly `talk' to the embeddings of different agents, different time steps, and different modes. In \cref{subsec:Agent Attention}, \cref{subsec:Historical Prediction Attention}, and \cref{subsec:Mode Attention}, we introduce Agent Attention, Historical Prediction Attention, and Mode Attention, respectively.

\subsection{Agent Attention}
\label{subsec:Agent Attention}

In the Agent Attention module,  self-attention is accomplished across agents for each mode and each time step on these prediction embeddings:
\begin{equation}
  P_{t,n,k}^A = \mathrm{MHA}(P_{t,n,k}, [P_{t,n',k}, E_e], [P_{t,n',k}, E_e]),
  \label{eq:agent attention}
\end{equation}
where \(n'\) denotes all agents within a certain radius \(R_2\) of the \(n\)-th agent under the same mode and time step. On the one hand, agent attention models the interactions among agents within their respective spatio-temporal contexts. On the other hand, it can also be conceived as the interaction between the future trajectories of different agents, thereby mitigating potential collisions.

\subsection{Historical Prediction Attention}
\label{subsec:Historical Prediction Attention}
After aggregating historical agent states, agent-lane interactions, and agent-agent interactions, previous methods typically begin predicting future trajectories. However, we observe that the current and historical predictions are usually correlated, while most existing methods neglect this. For example, when an agent is moving steadily on a straight path, the overlapping segments of successive predictions should be almost the same or vary minimally. When an agent traverses a busy multi-lane crossroads, the successive predictions may be quite different but still share the same motion goal as shown in \cref{fig:w/o HPA and w/ HPA} (b). Our experiments show that this kind of correlation between successive predictions is critical not only for the stability of the predictions but also for accuracy. 

Consequently, to further improve the stability and accuracy of trajectory prediction, we design this novel Historical Prediction Attention. It incorporates historical predictions to inform the current forecast, modeling the dynamic correlation between successive predictions by the attention mechanism. Specifically, each prediction embedding performs self-attention with historical prediction embeddings within the temporal span \(I_2\) for each agent and each mode:
\begin{equation}
  P_{t,n,k}^{HP} = \mathrm{MHA}(P_{t,n,k}^A, [P_{t-I_2\sim t,n,k}^A, E_e], [P_{t-I_2\sim t,n,k}^A, E_e]).
  \label{eq:historical prediction attention}
\end{equation}

Here, the prediction embeddings rather than final historical prediction trajectories are used to model the dynamic relationship, because the latter will change the training process from parallel execution to serial execution, greatly increasing the time required for training. 

Besides improving the accuracy and stability of prediction, this attention in Eq. (\ref{eq:historical prediction attention}) can also absorb longer history information beyond the currently visible window, \textit{i.e.}, extends the attention range. 
To be specific, without Historical Prediction Attention, the observation window of \(P_{t}^{HP}\) is confined to the interval \([t-I_1, t]\) as it only uses the \(I_1\) previous frames, limited to the time span \(I_1\). In contrast, if $I_2=I_1$, the observation window of Historical Prediction Attention is twice as long, \textit{i.e.},  \([t-I_1-I_1, t]\). In detail, the current prediction uses the historical $I_1$ prediction embedding, and thus the observation window is  \([t-I_1, t]\) w.r.t. the prediction embedding. However, the prediction embedding at the farthest timestep $t-I_1$ actually already absorbs attention information from previous \([t-I_1-I_1, t-I_1]\) frames.  Therefore, the actual observation window of Historical Prediction Attention is the sum of the two intervals, \textit{i.e.}, \([t-I_1-I_1, t]\). In general case that  $I_2 \neq I_1$, the actual observation window of Historical Prediction Attention is \([t-I_2-I_1, t]\), which is also longer than that of most existing methods \([t-I_1, t]\). \textit{The longer attention range of Historical Prediction Attention can provide more beneficial information for better trajectory prediction without additional computational cost.}

\subsection{Mode Attention}
\label{subsec:Mode Attention}
After the historical prediction attention, self-attention is then applied across different modes of prediction embeddings for each agent and each time step, modeling mode-mode interactions between different future trajectories to enhance multimodal outputs:
\begin{equation}
  P_{t,n,k}^{M} = \mathrm{MHA}(P_{t,n,k}^{HP}, [P_{t,n,1 \sim K}^{HP}, E_e], [P_{t,n,1 \sim K}^{HP}, E_e]).
  \label{eq:mode attention}
\end{equation}

After \cref{eq:mode attention}, the Triple Factorized Attention is accomplished, inducing enhanced prediction embedding. The Triple Factorized Attention is repeated \(N_{attn}=2\) times so that all prediction embeddings can fully interact with each other for more accurate prediction.

\subsection{Multimodal Output}
\label{subsec:Multimodal Output}
Finally, all the prediction embeddings are decoded through a two-layer MLP to obtain multiple future locations:
\begin{equation}
  L_{t,n,k}^1 = \mathrm{MLP}(P_{t,n,k}^{M}),
  \label{eq:decode trajectory}
\end{equation}
where \(L_{t,n,k}^1\in \mathbb{R}^{F \times 2}\).
To further enhance the output trajectories, following QCNet \cite{zhou2023query}, \(L_{t,n,k}^1\) is taken as the input of the whole pipeline to further refine the predicted trajectories. In detail, $L_{t,n,k}^1$ is taken as trajectory proposals and encoded into mode queries by another two-layer MLP. These encoded mode queries replace the learnable mode queries as inputs of Spatio-Temporal Attention to re-aggregate the spatio-temporal context and perform Triple Factorized Attention again. This refinement process produces a trajectory refinement \(\Delta L_{t,n,k}\) and probability scores \(\hat{\pi}_{t,n,k}\). 

Then, the final predicted trajectory is obtained by summing the trajectory proposal and the trajectory refinement:
\begin{equation}
  L_{t,n,k}^2 = L_{t,n,k}^1 + \Delta L_{t,n,k}.
  \label{eq:final trajectory}
\end{equation}

\subsection{Training Objective}
\label{Training Objective}
Following the existing works \cite{nayakanti2023wayformer, liu2021multimodal, wang2023prophnet, zhou2023query, zhou2022hivt, gao2020vectornet, cui2019multimodal}, we adopt the winner-takes-all \cite{lee2016stochastic} strategy to optimize our model. For marginal prediction, the \(k_{t,n}\)-th mode  to be optimized is determined based on the minimum endpoint displacement between the predicted trajectory \(\{L_{t,n,k}^1\}_{k\in [1,K]}\) and the ground truth \(G_{t,n}=\{g_{t+1,n}, g_{t+2,n},...,g_{t+F,n}\}\):
\begin{equation}
  k_{t,n} = \mathop{\arg\min}\limits_{k\in [1,K]}(l_{t+F,n,k}^1 - g_{t+F,n}).
  \label{eq:mode select}
\end{equation}
Then, the regression loss function contains two Huber losses for the trajectory proposals and the refined final trajectories, respectively:
\begin{equation}
  \mathcal{L}_{reg1}^{t,n} = \mathcal{L}_{Huber}(L_{t,n,k_{t,n}}^1, G_{t,n}),
  \label{eq:regression loss 1}
\end{equation}
\begin{equation}
  \mathcal{L}_{reg2}^{t,n} = \mathcal{L}_{Huber}(L_{t,n,k_{t,n}}^2, G_{t,n}).
  \label{eq:regression loss 2}
\end{equation}
Besides, the probability scores are optimized by using the cross-entropy loss function:
\begin{equation}
  \mathcal{L}_{cls}^{t,n} = \mathcal{L}_{CE}(\{\hat{\pi}_{t,n,k}\}_{k\in[1,K]}, k_{t,n}).
  \label{eq:classification loss}
\end{equation}
Overall, the total loss function of the whole model is formulated as follows:
\begin{equation}
  \mathcal{L} = \frac{1}{TN}\sum_{t=-T+1}^0\sum_{n=1}^N({L}_{reg1}^{t,n}+{L}_{reg2}^{t,n}+{L}_{cls}^{t,n}).
  \label{eq:total loss}
\end{equation}

For joint prediction, we treat the prediction of all agents in the same mode as a predicted future and the joint endpoint displacement determines the mode to be optimized. Please refer to the supplementary material for a detailed explanation of the training objective for joint prediction.

%% file: sec/4_experiment.tex
\section{Experiments}
\label{sec:experiments}

\begin{table*}
  \centering
  \begin{tabular}{c|l|cccc}
    \toprule
    & Method & \textbf{b-minFDE}\(\downarrow\) & minFDE\(\downarrow\) & MR\(\downarrow\) & minADE\(\downarrow\) \\
    \midrule
    \multirow{9}{*}{Single model} & LaneGCN \cite{liang2020learning} & 2.0539 & 1.3622 & 0.1620 & 0.8703\\
    & mmTransformer \cite{liu2021multimodal} & 2.0328 & 1.3383 & 0.1540 & 0.8436 \\
    & DenseTNT \cite{gu2021densetnt} & 1.9759 & 1.2815 & 0.1258 & 0.8817 \\
    & THOMAS \cite{gilles2022thomas} & 1.9736 & 1.4388 & 0.1038 & 0.9423 \\
    & TPCN \cite{ye2021tpcn} & 1.9286 & 1.2442 & 0.1333 & 0.8153 \\
    & SceneTransformer \cite{ngiam2021scene} & 1.8868 & 1.2321 & 0.1255 & 0.8026 \\
    & HiVT \cite{zhou2022hivt} & 1.8422 & 1.1693 & 0.1267 & \underline{0.7735} \\
    & GANet \cite{wang2023ganet} & \underline{1.7899} & \underline{1.1605} & \underline{0.1179} & 0.8060 \\
    & \textbf{HPNet(single model)} & \textbf{1.7375} & \textbf{1.0986} & \textbf{0.1067} & \textbf{0.7612} \\
    \midrule
    \multirow{9}{*}{Ensembled model} & HOME+GOHOME \cite{gilles2021home, gilles2022gohome} & 1.8601 & 1.2919 & \textbf{0.0846} & 0.8904 \\
    & Multipath++ \cite{varadarajan2022multipath++} & 1.7932 & 1.2144 & 0.1324 & 0.7897 \\
    & Macformer \cite{feng2023macformer} & 1.7667 & 1.2141 & 0.1272 & 0.8121 \\
    & DCMS \cite{ye2022dcms} & 1.7564 & 1.1350 & 0.1094 & 0.7659 \\
    & HeteroGCN \cite{gao2023dynamic} & 1.7512 & 1.1602 & 0.1168 & 0.7890 \\
    & Wayformer \cite{nayakanti2023wayformer} & 1.7408 & 1.1616 & 0.1186 & 0.7676 \\
    & ProphNet \cite{wang2023prophnet} & 1.6942 & 1.1337 & 0.1101 & 0.7623 \\
    & QCNet \cite{zhou2023query} & \underline{1.6934} & \textbf{1.0666} & \underline{0.1056} & \textbf{0.7340} \\
    & \textbf{HPNet(ensembled model)} & \textbf{1.6768} & \underline{1.0856} & 0.1075 & \underline{0.7478} \\
    \bottomrule
  \end{tabular}
  \caption{Comparison of HPNet with the state-of-the-art methods on the Argoverse test set, where b-minFDE is the official ranking metric.  For each metric, the best result is in \textbf{bold} and the second best result is \underline{underlined}.}
  \label{tab: Argoverse leaderboard}
\end{table*}

\begin{table}
  \centering
  \begin{tabular}{l|cc}
    \toprule
    Method & minJointFDE\(\downarrow\) & minJointADE\(\downarrow\) \\
    \midrule
    AutoBot \cite{girgis2021latent} & 1.0148 & 0.3123\\
    THOMAS \cite{gilles2022thomas} & 0.9679 & 0.4164\\
    Trai-MAE \cite{chen2023traj} & 0.9660 & 0.3066 \\
    HDGT \cite{jia2023hdgt} & 0.9580 & 0.3030 \\
    FJMP \cite{rowe2023fjmp} & \underline{0.9218} & \underline{0.2752} \\
    \midrule
    \textbf{HPNet(single model)} & \textbf{0.8231} & \textbf{0.2548} \\
    \bottomrule
  \end{tabular}
  \caption{Comparison of HPNet with the state-of-the-art methods on the INTERACTION test set. For each metric, the best result is in \textbf{bold} and the second best result is \underline{underlined}.}
  \label{tab: INTERACTION leaderboard}
\end{table}

\subsection{Experimental Setup}
\textbf{Dataset.}
We conduct the experiments on the Argoverse \cite{chang2019argoverse} and INTERACTION \cite{zhan2019interaction} datasets. Both are based on real-world driving scenarios, providing high-definition maps and detailed motion information, sampled at a frequency of 10Hz. On the Argoverse dataset, we assessed HPNet's capability for marginal trajectory prediction. Conversely, on the INTERACTION dataset, renowned for its complex driving scenarios and detailed multi-agent interactions, we examined HPNet's effectiveness in joint prediction.

\textbf{Metrics.}
For our evaluation, we employed official trajectory forecasting metrics, encompassing minimum Average Displacement Error (minADE), minimum Final Displacement Error (minFDE), Miss Rate (MR), and Brier-minimum Final Displacement Error (b-minFDE) for Argoverse. MinADE measures the average \( \ell_2 \)-norm distance across predicted and actual trajectory points, while minFDE examines the \( \ell_2 \)-norm distance at the trajectory's endpoint. MR assesses instances where predictions stray more than 2.0 meters from the actual endpoint, gauging model reliability. Lastly, brier-minFDE extends minFDE by incorporating the probability part \((1-\hat{\pi})^2\), providing insights into the model's confidence in its best prediction. For the INTERACTION dataset, we employed minJointADE and minJointFDE metrics to assess joint trajectory prediction performance. MinJointADE evaluates the average \( \ell_2 \)-norm distance across all agents' predicted and actual trajectories, while minJointFDE focuses on the \( \ell_2 \)-norm distance at the final time step for all agents. To explore the model's capability in capturing multimodal outputs, we set \( K = 6 \) for both marginal prediction and joint prediction.

\begin{table*}
    \centering
    \begin{tabular}{cc|ccc|cccc}
        \toprule
        \multicolumn{2}{c|}{Spatio-Temporal Attention} & \multicolumn{3}{c|}{Triple Factorized Attention} & \multicolumn{4}{c}{Metrics} \\
        \midrule
        Spatial & Temporal & Agent & Historical Prediction & Mode & b-minFDE\(\downarrow\) & minFDE\(\downarrow\) & MR\(\downarrow\) & minADE\(\downarrow\)\\
        \midrule
        $\checkmark$ & $\checkmark$ &  &  &  & 1.832 & 1.203 & 0.126 & 0.771 \\
        \midrule
        $\checkmark$ & $\checkmark$ &  & $\checkmark$ & $\checkmark$ & 1.711 & 1.084 & 0.102	& 0.722 \\
        $\checkmark$ & $\checkmark$ & $\checkmark$ &  & $\checkmark$ & 1.527 & 0.909 & 0.075 & 0.661 \\
        $\checkmark$ & $\checkmark$ & $\checkmark$ & $\checkmark$ &  & 1.531 & 0.894 & 0.073 & 0.645 \\
        \midrule
        $\checkmark$ & $\checkmark$ & $\checkmark$ & $\checkmark$ & $\checkmark$ & 1.506 & 0.871 & 0.069 & 0.638 \\
        \bottomrule
    \end{tabular}
    \caption{Ablation study of Triple Factorized Attention. Experiments are performed on the Argoverse validation set.}
    \label{tab: Ablation Study of Triple Factorized Attention}
\end{table*}

\subsection{Comparison with State-of-the-art}
\textbf{Results on Argoverse.} 
The marginal trajectory prediction results on
 Argoverse are reported in \cref{tab: Argoverse leaderboard}. Our HPNet achieves the best results on all indicators among all single models. Compared to GANet in second place, the improvement is up to 0.052 in b-minFDE, 0.062 in minFDE, and 0.045 in minADE.  Moreover, following \cite{varadarajan2022multipath++, ye2022dcms,  zhou2023query, wang2023prophnet}, HPNet is further compared in the setting of model ensembling. As can be seen, our HPNet also performs best in the official ranking metric. Compared to the single model, ensembled HPNet only yields a reduction of 0.013 in minFDE. This is mainly because the HPNet's predictions are more stable, and thus the improvement from ensembling is smaller than other methods. Overall, our HPNet achieves state-of-the-art performance, verifying the superiority of our HPNet. 

\textbf{Results on INTERACTION.} 
\cref{tab: INTERACTION leaderboard} shows the results of our method on the INTERACTION multi-agent track. We achieved state-of-the-art performance on this benchmark, achieving substantial gains over the second-ranked FJMP, with improvements of 0.099 in minJointFDE and 0.020 in minJointADE. This indicates that our HPNet can be used simply and effectively for joint trajectory prediction.

\subsection{Ablation Study}
\label{Ablation Study}
We first conduct ablation studies on Triple Factorized Attention to analyze the importance of Agent Attention, Historical Prediction Attention, and Mode Attention in our proposed HPNet. Then we explore the impact of Historical Prediction Attention on prediction accuracy and stability. Lastly, we examine the influence of Historical Prediction Attention on the 
reaction timeliness.

\textbf{Component Study of Triple Factorized Attention.}
As shown in \cref{tab: Ablation Study of Triple Factorized Attention}, the model with all components achieves 1.506 in terms of b-minFDE, which is the best result on the validation set. If removing Triple Factorized Attention, the performance in terms of b-minFDE drops by 0.326, implying the importance of the Triple Factorized Attention module in the overall model architecture. If removing the Agent Attention, Historical Prediction Attention, and Mode Attention, the performance in terms of b-minFDE drops by 0.205, 0.021, and 0.025, respectively. This indicates the effectiveness of all three attention modules, among which the Agent Attention between agents and surrounding agents has the most important influence and is indispensable for prediction. Besides, our proposed Historical Prediction Attention also plays an important role with obvious improvements on four metrics, which clearly illustrates the necessity of considering the relationship between successive predictions.

\begin{figure}[t]
  \centering
  \includegraphics[width=1.0\linewidth]{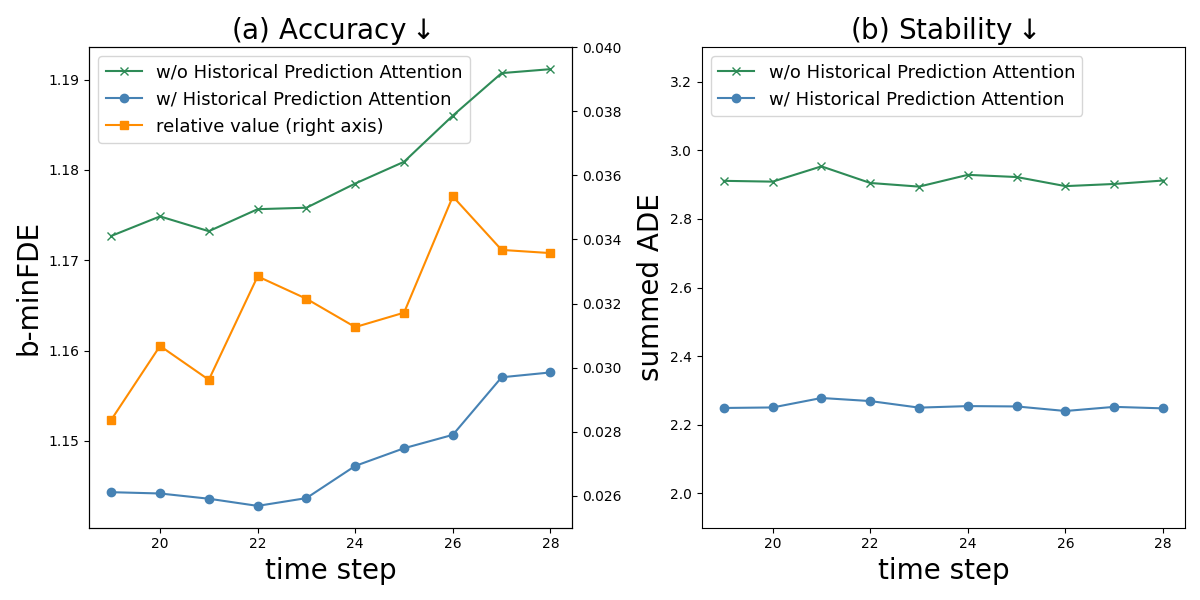}
   \caption{Comparison of prediction Accuracy (b-minFDE\(\downarrow\)) and Stability (summed ADE\(\downarrow\)) of our HPNet and its baseline without Historical Prediction Attention on the Argoverse validation set.}
   \label{fig: accuracy and stability}
\end{figure}

\begin{figure*}
  \centering
  \begin{subfigure}{0.24\linewidth}
    \includegraphics[width=1.0\linewidth]{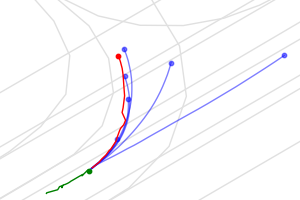}
  \end{subfigure}
  \hfill
  \begin{subfigure}{0.24\linewidth}
    \includegraphics[width=1.0\linewidth]{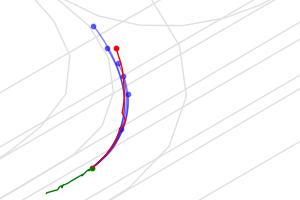}
  \end{subfigure}
  \hfill
  \begin{subfigure}{0.24\linewidth}
    \includegraphics[width=1.0\linewidth]{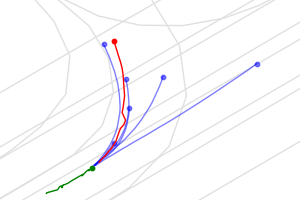}
  \end{subfigure}
  \hfill
  \begin{subfigure}{0.24\linewidth}
    \includegraphics[width=1.0\linewidth]{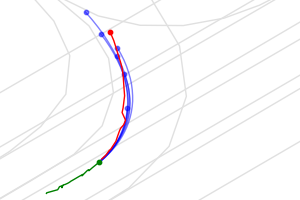}
  \end{subfigure}
  
  \centering
  {\small (a) baseline (w/o Historical Prediction Attention) in 20-24 four time steps.}

  \begin{subfigure}{0.24\linewidth}
    \includegraphics[width=1.0\linewidth]{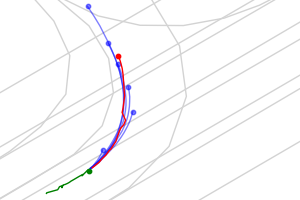}
  \end{subfigure}
  \hfill
  \begin{subfigure}{0.24\linewidth}
    \includegraphics[width=1.0\linewidth]{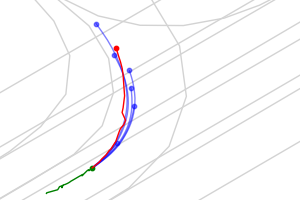}
  \end{subfigure}
  \hfill
  \begin{subfigure}{0.24\linewidth}
    \includegraphics[width=1.0\linewidth]{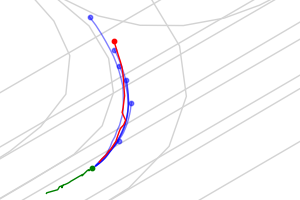}
  \end{subfigure}
  \hfill
  \begin{subfigure}{0.24\linewidth}
    \includegraphics[width=1.0\linewidth]{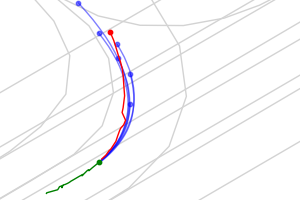}
  \end{subfigure}
  
  \centering
  {\small (b) HPNet (w/ Historical Prediction Attention) in 20-24 four time steps.}
  
  \caption{Qualitative results on the Argoverse validation set. Baseline (a) alternately forecasts one motion goal (\textit{i.e.}, turn left) and two motion goals (\textit{i.e.}, turn left and go straight). In contrast, HPNet (b) consistently and reliably predicts the same motion goal (\textit{i.e.}, turn left). The lanes, historical trajectory, ground truth trajectory, and six predicted trajectories are indicated in grey, green, red, and blue, respectively.}
  \label{fig:w/o HPA and w/ HPA}
\end{figure*}

\begin{figure}[t]
    \centering
    \begin{subfigure}[b]{0.32\columnwidth}
        \includegraphics[width=\linewidth]{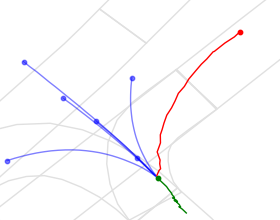}
    \end{subfigure}
    \hfill
    \begin{subfigure}[b]{0.32\columnwidth}
        \includegraphics[width=\linewidth]{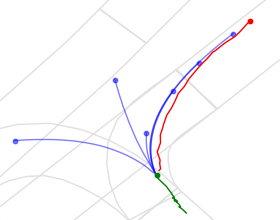}
    \end{subfigure}
    \hfill
    \begin{subfigure}[b]{0.32\columnwidth}
        \includegraphics[width=\linewidth]{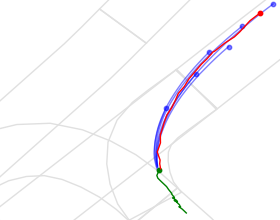}
    \end{subfigure}

    \begin{subfigure}[b]{0.32\columnwidth}
        \includegraphics[width=\linewidth]{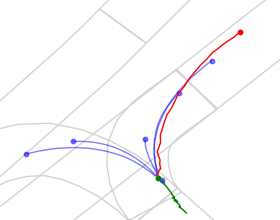}
    \end{subfigure}
    \hfill
    \begin{subfigure}[b]{0.32\columnwidth}
        \includegraphics[width=\linewidth]{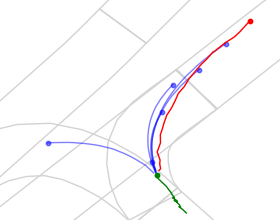}
    \end{subfigure}
    \hfill
    \begin{subfigure}[b]{0.32\columnwidth}
        \includegraphics[width=\linewidth]{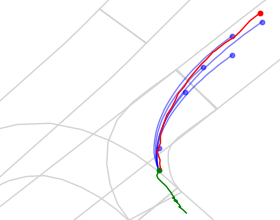}
    \end{subfigure}

    \caption{Predictions of HPNet (lower) and baseline (upper). }
    \label{fig: timeliness of reaction}
\end{figure}

\textbf{The Impact of Historical Prediction Attention on Accuracy and Stability.}
Our proposed Historical Prediction Attention is expected to improve the accuracy and stability of trajectory prediction by considering the relationship between current and historical predictions. To investigate whether this expectation is achieved, we conduct comparative experiments on two models:  the HPNet and its baseline model without Historical Prediction Attention. Predictions are made across 10 time steps, ranging from 20 to 30, with each prediction utilizing a visible history flame window of 20 time steps and a historical prediction window of equal length.   The accuracy of the predictions is quantified by using the b-minFDE metric.   The stability is assessed by the summed ADE of the overlapping segments of matched trajectory pairs at current and previous time steps, where the matched trajectory pairs are obtained via the Hungarian matching algorithm. 

As shown in \cref{fig: accuracy and stability} (a), for all predicted time steps, the performance of HPNet in terms of b-minFDE is better than that of the baseline. This superior performance indicates that Historical Prediction Attention indeed improves the accuracy of trajectory prediction. Besides, it is also observed that the accuracy of both our HPNet and the baseline declines along the temporal axis, and this is mainly because of the appearance of new agents in later time steps that are not present in the first 20 frames. Even so, the relative improvement of our HPNet over the baseline (\textit{i.e.}, the orange dashed line) becomes larger over time. When the time step becomes longer, a big difference is that the actual visible historical window of our HPNet is beyond 20 time steps as analyzed in \cref{subsec:Historical Prediction Attention}, while that of the baseline is always fixed to 20 time steps. Therefore, this significant relative improvement over time clearly verifies the benefit from the longer attention window of Historical Prediction Attention. 

As shown in \cref{fig: accuracy and stability} (b), for all predicted time steps, the summed ADE of HPNet between successive time steps is about 2.25, while the summed ADE of the baseline is about 2.90 which is much larger. This indicates that Historical Prediction Attention indeed makes predicted trajectories much more stable. We show an example in \cref{fig:w/o HPA and w/ HPA} to make the comparison more intuitive. As shown, the agent chooses to turn left at the intersection. However, the baseline may be due to the agent's pause in the middle moment, alternately predicting one motion goal (\textit{i.e.}, turn left) and two motion goals (\textit{i.e.}, turn left and go straight). In contrast, HPNet consistently and reliably predicted the same motion goal (\textit{i.e.}, turn left). At the same time, unlike DCMS \cite{ye2022dcms}, the successive predictions of HPNet only share the same motion goal in complex road conditions, without forced overlapping waypoints. These stable prediction results enable subsequent modules to produce stable and time-consistent safe driving decisions. More qualitative results can be found in the supplementary material.

\textbf{The Influence of Historical Prediction Attention on Reaction Timeliness.}
While Historical Prediction Attention enhances forecasting stability by leveraging historical predictions, does it hurt the reaction timeliness? Our answer is no. This is mainly benefited from the attention mechanism. When an abrupt change occurs (\textit{e.g.}, a sudden right turn in \cref{fig: timeliness of reaction}), the similarity between the current and historical prediction embeddings decreases, leading to reduced weights for historical predictions. Consequently, the impact of past predictions on the current moment dynamically diminishes. \cref{fig: timeliness of reaction} illustrates a qualitative example where an agent at an intersection is observed at three consecutive moments. When the agent shows no specific intention, HPNet stably and accurately forecasts the possibilities of turning left or right, outperforming the baseline. At the sudden right turn (in the final moment), HPNet quickly adjusts to predict right turns only, with no delay compared to the baseline.

%% file: sec/5_conclusion.tex
\section{Conclusion}
\label{sec:conclusion}
In this paper, we propose a novel dynamic trajectory prediction method, HPNet. A Historical Prediction Attention module is designed to model the dynamic relationship between successive predictions. It employs historical prediction embeddings to guide current forecast, making the predicted trajectories more accurate and stable. Experiments on the Argoverse and INTERACTION datasets demonstrate that our proposed HPNet achieves state-of-the-art performance, and also proves that Historical Prediction Attention can effectively improve accuracy and stability. 

\textbf{Acknowledgment.}
This work was partially supported by the National Key R\&D Program of China (No. 2020AAA0104500) and the Natural Science Foundation of China (Nos. U2336213 and 62122074).

%% file: sec/X_suppl.tex
\clearpage
\setcounter{page}{1}
\maketitlesupplementary

\label{sec:quality results}
\begin{figure*}
  \centering
  \begin{subfigure}{0.24\linewidth}
    \includegraphics[width=1.0\linewidth]{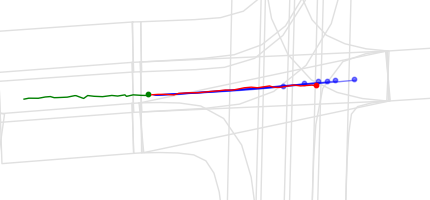}
  \end{subfigure}
  \hfill
  \begin{subfigure}{0.24\linewidth}
    \includegraphics[width=1.0\linewidth]{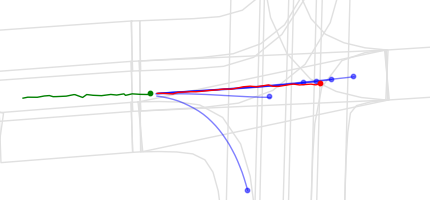}
  \end{subfigure}
  \hfill
  \begin{subfigure}{0.24\linewidth}
    \includegraphics[width=1.0\linewidth]{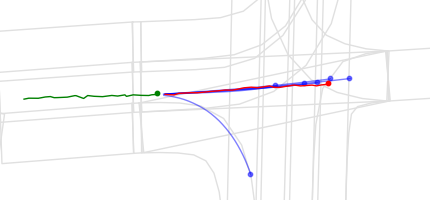}
  \end{subfigure}
  \hfill
  \begin{subfigure}{0.24\linewidth}
    \includegraphics[width=1.0\linewidth]{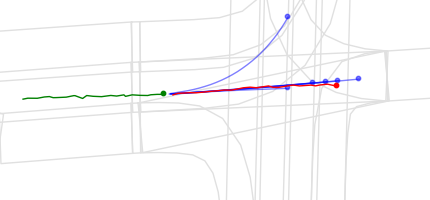}
  \end{subfigure}

  \begin{subfigure}{0.24\linewidth}
    \includegraphics[width=1.0\linewidth]{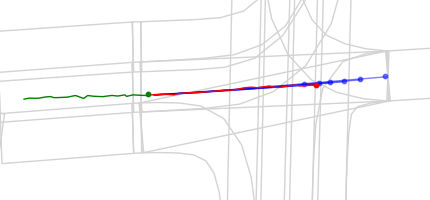}
  \end{subfigure}
  \hfill
  \begin{subfigure}{0.24\linewidth}
    \includegraphics[width=1.0\linewidth]{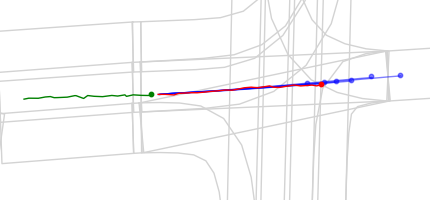}
  \end{subfigure}
  \hfill
  \begin{subfigure}{0.24\linewidth}
    \includegraphics[width=1.0\linewidth]{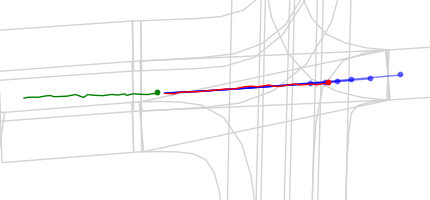}
  \end{subfigure}
  \hfill
  \begin{subfigure}{0.24\linewidth}
    \includegraphics[width=1.0\linewidth]{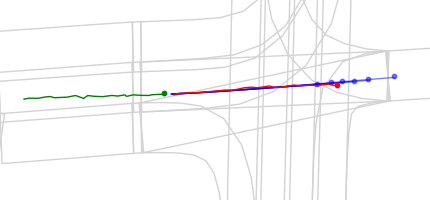}
  \end{subfigure}

  \centering
  {\small (a) baseline (w/o Historical Prediction Attention, upper) and HPNet (w/ Historical Prediction Attention, lower) in 20-24 four time steps.}

  \begin{subfigure}{0.24\linewidth}
    \includegraphics[width=1.0\linewidth]{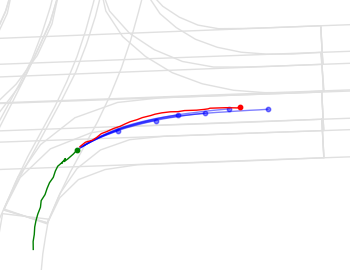}
  \end{subfigure}
  \hfill
  \begin{subfigure}{0.24\linewidth}
    \includegraphics[width=1.0\linewidth]{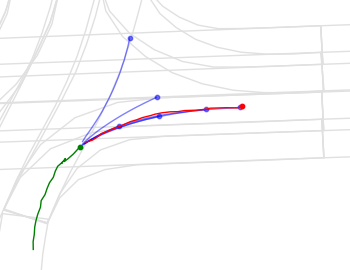}
  \end{subfigure}
  \hfill
  \begin{subfigure}{0.24\linewidth}
    \includegraphics[width=1.0\linewidth]{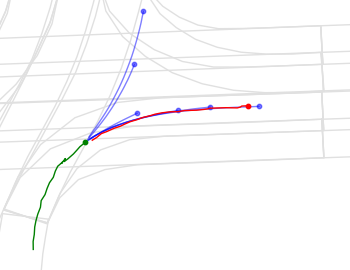}
  \end{subfigure}
  \hfill
  \begin{subfigure}{0.24\linewidth}
    \includegraphics[width=1.0\linewidth]{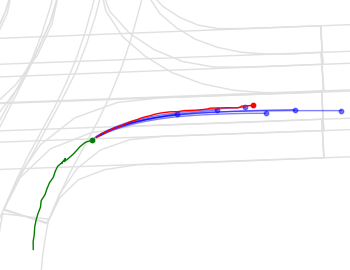}
  \end{subfigure}

  \begin{subfigure}{0.24\linewidth}
    \includegraphics[width=1.0\linewidth]{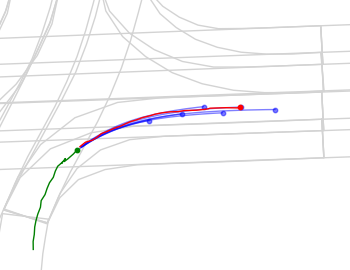}
  \end{subfigure}
  \hfill
  \begin{subfigure}{0.24\linewidth}
    \includegraphics[width=1.0\linewidth]{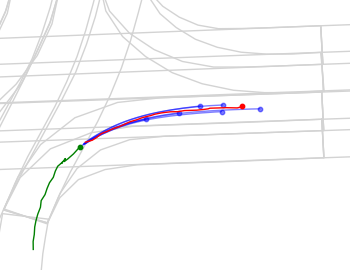}
  \end{subfigure}
  \hfill
  \begin{subfigure}{0.24\linewidth}
    \includegraphics[width=1.0\linewidth]{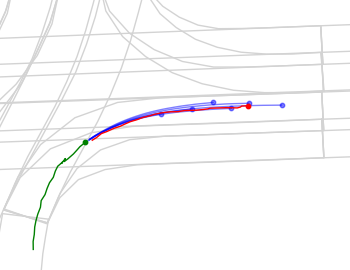}
  \end{subfigure}
  \hfill
  \begin{subfigure}{0.24\linewidth}
    \includegraphics[width=1.0\linewidth]{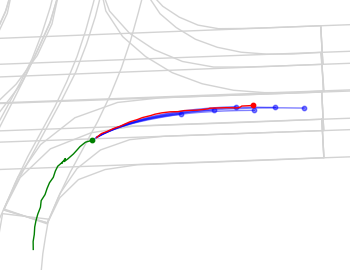}
  \end{subfigure}

  \centering
  {\small (b) baseline (w/o Historical Prediction Attention, upper) and HPNet (w/ Historical Prediction Attention, lower) in 20-24 four time steps.}
  
  \caption{Qualitative results on the Argoverse validation set. The lanes, historical trajectory, ground truth trajectory, and six predicted trajectories are indicated in grey, green, red, and blue, respectively.}
  \label{fig: qualitative results}
\end{figure*}

\section{Training Objective for Joint Prediction}
With a simple adjustment of the training objective, HPNet can be used for joint trajectory prediction. In joint prediction, we treat the predictions of all agents in the same mode as a single predicted future, so the \(k_{t}\)-th mode to be optimized is determined based on the minimum joint endpoint displacements between the predicted future \(\{L_{t,1\sim N,k}^1\}_{k\in [1,K]}\) and the ground truth \(G_{t, 1\sim N} = \{g_{t+1,1}, g_{t+1,2},..., g_{t+F,N}\}\):
\begin{equation}
    k_{t} = \mathop{\arg\min}\limits_{k\in [1,K]}\sum_{n=1}^N(l_{t+F,n,k}^1 - g_{t+F,n})
\end{equation}
Then, the regression loss function contains two Huber losses for the trajectory proposals and the refined final trajectories, respectively:
\begin{equation}
  \mathcal{L}_{reg1}^{t} = \sum_{n=1}^N\mathcal{L}_{Huber}(L_{t,n,k_{t}}^1, G_{t,n}),
  \label{eq: regression loss 1}
\end{equation}
\begin{equation}
  \mathcal{L}_{reg2}^{t} = \sum_{n=1}^N\mathcal{L}_{Huber}(L_{t,n,k_{t}}^2, G_{t,n}).
  \label{eq: regression loss 2}
\end{equation}
Besides, the future probability scores \(\hat{\pi}_{t,k}\) are optimized by using the cross-entropy loss function:
\begin{equation}
  \mathcal{L}_{cls}^{t} = \mathcal{L}_{CE}(\{\hat{\pi}_{t,k}\}_{k\in[1,K]}, k_{t}).
  \label{eq: classification loss}
\end{equation}
Overall, the total loss function of the whole model is formulated as follows:
\begin{equation}
  \mathcal{L} = \frac{1}{T}\sum_{t=-T+1}^0({L}_{reg1}^{t}+{L}_{reg2}^{t}+{L}_{cls}^{t}).
  \label{eq: total loss}
\end{equation}

\section{Implementation Details}
Our model trains on 8 RTX 4090 GPUs for 64 epochs, using the AdamW \cite{loshchilov2018decoupled} optimizer with a batch size of 16, dropout rate of 0.1, and weight decay of \( 1\times 10^{-4} \). Initial learning rates are \( 5\times 10^{-4} \)  for Argoverse and \( 3\times 10^{-4} \) for INTERACTION, with a cosine annealing scheduler for rate decay. On Argoverse, we apply a single Spatio-Temporal Attention layer and two Triple Factorized Attention layers, setting a 50 radius for all local areas and a 20-time span for historical frames and predictions. Data augmentation techniques—horizontal flipping, agent occlusion, and lane occlusion—are used with ratios of 0.5, 0.05, and 0.2, respectively. On INTERACTION, the setup involves one Spatio-Temporal Attention layer and three Triple Factorized Attention layers, with an 80 radius for local areas and a 10-time span for frames and predictions. The augmentation techniques of horizontal flipping and lane occlusion remain the same without the use of agent occlusion. The model sizes for Argoverse and INTERACTION are 4.1M and 5.3M,respectively.

\section{Inference Latency}
We also take a closer look at the inference latency, which is important for practical applications. The inference latency is reported on the Argoverse validation set with an NVIDIA V100 GPU. In the training stage, parallel processing is employed for fast training. In the testing stage, both parallel processing and serial processing can be used depending on the practical requirements. With serial processing, the time required for HPNet to predict all trajectories for all agents in a single time step is 27.62 ms, and the baseline model without Historical Prediction Attention takes 22.76 ms. When parallel processing, the time required for HPNet and the baseline model to predict trajectories for all agents at all 20 time steps is 92.02 ms and 81.08 ms respectively. These results indicate that the inference latency introduced by Historical Prediction Attention is small. In practice, we suggest serial processing. Overall, Historical Prediction Attention will not impede the real-time operational capabilities of autonomous driving systems, while concurrently enhancing prediction performance.

\section{Quality Results}
In \cref{fig: qualitative results}, we show two representative examples. 
In \cref{fig: qualitative results} (a), the agent is in the middle lane and is about to move forward. The baseline gives three types of possible motion goals (\textit{i.e.}, go straight, turn right and go straight, turn left and go straight) at four consecutive moments, which undoubtedly complicates the subsequent decision-making process. In contrast, HPNet provides a stable, accurate motion goal (\textit{i.e.}, go straight).
In \cref{fig: qualitative results} (b), the agent exhibits a clear intention to turn right. The baseline gave two motion goals at the middle two moments (\textit{i.e.}, go straight and turn right), while at each of the other two moments, it gave only one motion goal (\textit{i.e.}, turn right).In contrast, HPNet also gives a stable and accurate motion goal (\textit{i.e.}, turn right).
All in all, these examples clearly and intuitively demonstrate the great improvement of Historical Prediction Attention on the accuracy and stability of trajectory prediction, indicating its importance and effectiveness.

%% file: main.bbl
\begin{thebibliography}{48}
\providecommand{\natexlab}[1]{#1}
\providecommand{\url}[1]{\texttt{#1}}
\expandafter\ifx\csname urlstyle\endcsname\relax
  \providecommand{\doi}[1]{doi: #1}\else
  \providecommand{\doi}{doi: \begingroup \urlstyle{rm}\Url}\fi

\bibitem[Carion et~al.(2020)Carion, Massa, Synnaeve, Usunier, Kirillov, and Zagoruyko]{carion2020end}
Nicolas Carion, Francisco Massa, Gabriel Synnaeve, Nicolas Usunier, Alexander Kirillov, and Sergey Zagoruyko.
\newblock End-to-end object detection with transformers.
\newblock In \emph{Proceedings of the European Conference on Computer Vision (ECCV)}, 2020.

\bibitem[Casas et~al.(2020)Casas, Gulino, Suo, Luo, Liao, and Urtasun]{casas2020implicit}
Sergio Casas, Cole Gulino, Simon Suo, Katie Luo, Renjie Liao, and Raquel Urtasun.
\newblock Implicit latent variable model for scene-consistent motion forecasting.
\newblock In \emph{Proceedings of the European Conference on Computer Vision (ECCV)}, 2020.

\bibitem[Chai et~al.(2020)Chai, Sapp, Bansal, and Anguelov]{chai2020multipath}
Yuning Chai, Benjamin Sapp, Mayank Bansal, and Dragomir Anguelov.
\newblock Multipath: Multiple probabilistic anchor trajectory hypotheses for behavior prediction.
\newblock In \emph{Conference on Robot Learning (CoRL)}, 2020.

\bibitem[Chang et~al.(2019)Chang, Lambert, Sangkloy, Singh, Bak, Hartnett, Wang, Carr, Lucey, Ramanan, et~al.]{chang2019argoverse}
Ming-Fang Chang, John Lambert, Patsorn Sangkloy, Jagjeet Singh, Slawomir Bak, Andrew Hartnett, De Wang, Peter Carr, Simon Lucey, Deva Ramanan, et~al.
\newblock Argoverse: 3d tracking and forecasting with rich maps.
\newblock In \emph{Proceedings of the IEEE/CVF Conference on Computer Vision and Pattern Recognition (CVPR)}, 2019.

\bibitem[Chen et~al.(2023)Chen, Wang, Shao, Liu, Hao, Guan, Chen, and Heng]{chen2023traj}
Hao Chen, Jiaze Wang, Kun Shao, Furui Liu, Jianye Hao, Chenyong Guan, Guangyong Chen, and Pheng-Ann Heng.
\newblock Traj-mae: Masked autoencoders for trajectory prediction.
\newblock In \emph{Proceedings of the IEEE/CVF International Conference on Computer Vision (ICCV)}, 2023.

\bibitem[Cheng et~al.(2023)Cheng, Mei, and Liu]{cheng2023forecast}
Jie Cheng, Xiaodong Mei, and Ming Liu.
\newblock Forecast-mae: Self-supervised pre-training for motion forecasting with masked autoencoders.
\newblock In \emph{Proceedings of the IEEE/CVF International Conference on Computer Vision (ICCV)}, 2023.

\bibitem[Choi et~al.(2023)Choi, Kim, Yun, and Choi]{choi2023r}
Sehwan Choi, Jungho Kim, Junyong Yun, and Jun~Won Choi.
\newblock R-pred: Two-stage motion prediction via tube-query attention-based trajectory refinement.
\newblock In \emph{Proceedings of the IEEE/CVF International Conference on Computer Vision (ICCV)}, 2023.

\bibitem[Cui et~al.(2019)Cui, Radosavljevic, Chou, Lin, Nguyen, Huang, Schneider, and Djuric]{cui2019multimodal}
Henggang Cui, Vladan Radosavljevic, Fang-Chieh Chou, Tsung-Han Lin, Thi Nguyen, Tzu-Kuo Huang, Jeff Schneider, and Nemanja Djuric.
\newblock Multimodal trajectory predictions for autonomous driving using deep convolutional networks.
\newblock In \emph{IEEE International Conference on Robotics and Automation (ICRA)}, 2019.

\bibitem[Deo et~al.(2022)Deo, Wolff, and Beijbom]{deo2022multimodal}
Nachiket Deo, Eric Wolff, and Oscar Beijbom.
\newblock Multimodal trajectory prediction conditioned on lane-graph traversals.
\newblock In \emph{Conference on Robot Learning (CoRL)}, 2022.

\bibitem[Feng et~al.(2023)Feng, Zhou, Lin, Zhang, Xu, Zhang, Zhou, and Shen]{feng2023macformer}
Chen Feng, Hangning Zhou, Huadong Lin, Zhigang Zhang, Ziyao Xu, Chi Zhang, Boyu Zhou, and Shaojie Shen.
\newblock Macformer: Map-agent coupled transformer for real-time and robust trajectory prediction.
\newblock \emph{IEEE Robotics and Automation Letters (RA-L)}, 2023.

\bibitem[Gao et~al.(2020)Gao, Sun, Zhao, Shen, Anguelov, Li, and Schmid]{gao2020vectornet}
Jiyang Gao, Chen Sun, Hang Zhao, Yi Shen, Dragomir Anguelov, Congcong Li, and Cordelia Schmid.
\newblock Vectornet: Encoding hd maps and agent dynamics from vectorized representation.
\newblock In \emph{Proceedings of the IEEE/CVF Conference on Computer Vision and Pattern Recognition (CVPR)}, 2020.

\bibitem[Gao et~al.(2023)Gao, Jia, Li, and Xiong]{gao2023dynamic}
Xing Gao, Xiaogang Jia, Yikang Li, and Hongkai Xiong.
\newblock Dynamic scenario representation learning for motion forecasting with heterogeneous graph convolutional recurrent networks.
\newblock \emph{IEEE Robotics and Automation Letters (RA-L)}, 2023.

\bibitem[Gilles et~al.(2021)Gilles, Sabatini, Tsishkou, Stanciulescu, and Moutarde]{gilles2021home}
Thomas Gilles, Stefano Sabatini, Dzmitry Tsishkou, Bogdan Stanciulescu, and Fabien Moutarde.
\newblock Home: Heatmap output for future motion estimation.
\newblock In \emph{IEEE International Intelligent Transportation Systems Conference (ITSC)}, 2021.

\bibitem[Gilles et~al.(2022{\natexlab{a}})Gilles, Sabatini, Tsishkou, Stanciulescu, and Moutarde]{gilles2022gohome}
Thomas Gilles, Stefano Sabatini, Dzmitry Tsishkou, Bogdan Stanciulescu, and Fabien Moutarde.
\newblock Gohome: Graph-oriented heatmap output for future motion estimation.
\newblock In \emph{IEEE International Conference on Robotics and Automation (ICRA)}, 2022{\natexlab{a}}.

\bibitem[Gilles et~al.(2022{\natexlab{b}})Gilles, Sabatini, Tsishkou, Stanciulescu, and Moutarde]{gilles2022thomas}
Thomas Gilles, Stefano Sabatini, Dzmitry Tsishkou, Bogdan Stanciulescu, and Fabien Moutarde.
\newblock Thomas: Trajectory heatmap output with learned multi-agent sampling.
\newblock In \emph{Proceedings of the International Conference on Learning Representations (ICLR)}, 2022{\natexlab{b}}.

\bibitem[Girgis et~al.(2021)Girgis, Golemo, Codevilla, Weiss, D'Souza, Kahou, Heide, and Pal]{girgis2021latent}
Roger Girgis, Florian Golemo, Felipe Codevilla, Martin Weiss, Jim~Aldon D'Souza, Samira~Ebrahimi Kahou, Felix Heide, and Christopher Pal.
\newblock Latent variable sequential set transformers for joint multi-agent motion prediction.
\newblock \emph{arXiv preprint arXiv:2104.00563}, 2021.

\bibitem[Gu et~al.(2021)Gu, Sun, and Zhao]{gu2021densetnt}
Junru Gu, Chen Sun, and Hang Zhao.
\newblock Densetnt: End-to-end trajectory prediction from dense goal sets.
\newblock In \emph{Proceedings of the IEEE/CVF International Conference on Computer Vision (ICCV)}, 2021.

\bibitem[Gupta et~al.(2018)Gupta, Johnson, Fei-Fei, Savarese, and Alahi]{gupta2018social}
Agrim Gupta, Justin Johnson, Li Fei-Fei, Silvio Savarese, and Alexandre Alahi.
\newblock Social gan: Socially acceptable trajectories with generative adversarial networks.
\newblock In \emph{Proceedings of the IEEE Conference on Computer Vision and Pattern Recognition (CVPR)}, 2018.

\bibitem[Jia et~al.(2023)Jia, Wu, Chen, Liu, Li, and Yan]{jia2023hdgt}
Xiaosong Jia, Penghao Wu, Li Chen, Yu Liu, Hongyang Li, and Junchi Yan.
\newblock Hdgt: Heterogeneous driving graph transformer for multi-agent trajectory prediction via scene encoding.
\newblock \emph{IEEE Transactions on Pattern Analysis and Machine Intelligence (TPAMI)}, 2023.

\bibitem[Kim et~al.(2021)Kim, Park, Lee, Khoshimjonov, Kum, Kim, Kim, and Choi]{kim2021lapred}
ByeoungDo Kim, Seong~Hyeon Park, Seokhwan Lee, Elbek Khoshimjonov, Dongsuk Kum, Junsoo Kim, Jeong~Soo Kim, and Jun~Won Choi.
\newblock Lapred: Lane-aware prediction of multi-modal future trajectories of dynamic agents.
\newblock In \emph{Proceedings of the IEEE/CVF Conference on Computer Vision and Pattern Recognition (CVPR)}, 2021.

\bibitem[Lee et~al.(2016)Lee, Purushwalkam Shiva~Prakash, Cogswell, Ranjan, Crandall, and Batra]{lee2016stochastic}
Stefan Lee, Senthil Purushwalkam Shiva~Prakash, Michael Cogswell, Viresh Ranjan, David Crandall, and Dhruv Batra.
\newblock Stochastic multiple choice learning for training diverse deep ensembles.
\newblock In \emph{Advances in Neural Information Processing Systems (NIPS)}, 2016.

\bibitem[Li et~al.(2023)Li, Dai, Zhu, Chen, Wang, and Xia]{li2023fsr}
Jinmin Li, Tao Dai, Mingyan Zhu, Bin Chen, Zhi Wang, and Shu-Tao Xia.
\newblock Fsr: A general frequency-oriented framework to accelerate image super-resolution networks.
\newblock In \emph{Proceedings of the AAAI Conference on Artificial Intelligence (AAAI)}, 2023.

\bibitem[Liang et~al.(2020)Liang, Yang, Hu, Chen, Liao, Feng, and Urtasun]{liang2020learning}
Ming Liang, Bin Yang, Rui Hu, Yun Chen, Renjie Liao, Song Feng, and Raquel Urtasun.
\newblock Learning lane graph representations for motion forecasting.
\newblock In \emph{Proceedings of the European Conference on Computer Vision (ECCV)}, 2020.

\bibitem[Liu et~al.(2021)Liu, Zhang, Fang, Jiang, and Zhou]{liu2021multimodal}
Yicheng Liu, Jinghuai Zhang, Liangji Fang, Qinhong Jiang, and Bolei Zhou.
\newblock Multimodal motion prediction with stacked transformers.
\newblock In \emph{Proceedings of the IEEE/CVF Conference on Computer Vision and Pattern Recognition (CVPR)}, 2021.

\bibitem[Loshchilov and Hutter(2018)]{loshchilov2018decoupled}
Ilya Loshchilov and Frank Hutter.
\newblock Decoupled weight decay regularization.
\newblock In \emph{Proceedings of the International Conference on Learning Representations (ICLR)}, 2018.

\bibitem[Nayakanti et~al.(2023)Nayakanti, Al-Rfou, Zhou, Goel, Refaat, and Sapp]{nayakanti2023wayformer}
Nigamaa Nayakanti, Rami Al-Rfou, Aurick Zhou, Kratarth Goel, Khaled~S Refaat, and Benjamin Sapp.
\newblock Wayformer: Motion forecasting via simple \& efficient attention networks.
\newblock In \emph{IEEE International Conference on Robotics and Automation (ICRA)}, 2023.

\bibitem[Ngiam et~al.(2021)Ngiam, Vasudevan, Caine, Zhang, Chiang, Ling, Roelofs, Bewley, Liu, Venugopal, et~al.]{ngiam2021scene}
Jiquan Ngiam, Vijay Vasudevan, Benjamin Caine, Zhengdong Zhang, Hao-Tien~Lewis Chiang, Jeffrey Ling, Rebecca Roelofs, Alex Bewley, Chenxi Liu, Ashish Venugopal, et~al.
\newblock Scene transformer: A unified architecture for predicting future trajectories of multiple agents.
\newblock In \emph{Proceedings of the International Conference on Learning Representations (ICLR)}, 2021.

\bibitem[Phan-Minh et~al.(2020)Phan-Minh, Grigore, Boulton, Beijbom, and Wolff]{phan2020covernet}
Tung Phan-Minh, Elena~Corina Grigore, Freddy~A Boulton, Oscar Beijbom, and Eric~M Wolff.
\newblock Covernet: Multimodal behavior prediction using trajectory sets.
\newblock In \emph{Proceedings of the IEEE/CVF Conference on Computer Vision and Pattern Recognition (CVPR)}, 2020.

\bibitem[Rowe et~al.(2023)Rowe, Ethier, Dykhne, and Czarnecki]{rowe2023fjmp}
Luke Rowe, Martin Ethier, Eli-Henry Dykhne, and Krzysztof Czarnecki.
\newblock Fjmp: Factorized joint multi-agent motion prediction over learned directed acyclic interaction graphs.
\newblock In \emph{Proceedings of the IEEE/CVF Conference on Computer Vision and Pattern Recognition (CVPR)}, 2023.

\bibitem[Sadeghian et~al.(2019)Sadeghian, Kosaraju, Sadeghian, Hirose, Rezatofighi, and Savarese]{sadeghian2019sophie}
Amir Sadeghian, Vineet Kosaraju, Ali Sadeghian, Noriaki Hirose, Hamid Rezatofighi, and Silvio Savarese.
\newblock Sophie: An attentive gan for predicting paths compliant to social and physical constraints.
\newblock In \emph{Proceedings of the IEEE/CVF Conference on Computer Vision and Pattern Recognition (CVPR)}, 2019.

\bibitem[Salzmann et~al.(2020)Salzmann, Ivanovic, Chakravarty, and Pavone]{salzmann2020trajectron++}
Tim Salzmann, Boris Ivanovic, Punarjay Chakravarty, and Marco Pavone.
\newblock Trajectron++: Dynamically-feasible trajectory forecasting with heterogeneous data.
\newblock In \emph{Proceedings of the European Conference on Computer Vision (ECCV)}, 2020.

\bibitem[Seff et~al.(2023)Seff, Cera, Chen, Ng, Zhou, Nayakanti, Refaat, Al-Rfou, and Sapp]{seff2023motionlm}
Ari Seff, Brian Cera, Dian Chen, Mason Ng, Aurick Zhou, Nigamaa Nayakanti, Khaled~S Refaat, Rami Al-Rfou, and Benjamin Sapp.
\newblock Motionlm: Multi-agent motion forecasting as language modeling.
\newblock In \emph{Proceedings of the IEEE/CVF International Conference on Computer Vision (ICCV)}, 2023.

\bibitem[Shi et~al.(2022)Shi, Jiang, Dai, and Schiele]{shi2022motion}
Shaoshuai Shi, Li Jiang, Dengxin Dai, and Bernt Schiele.
\newblock Motion transformer with global intention localization and local movement refinement.
\newblock \emph{Advances in Neural Information Processing Systems (NIPS)}, 2022.

\bibitem[Varadarajan et~al.(2022)Varadarajan, Hefny, Srivastava, Refaat, Nayakanti, Cornman, Chen, Douillard, Lam, Anguelov, et~al.]{varadarajan2022multipath++}
Balakrishnan Varadarajan, Ahmed Hefny, Avikalp Srivastava, Khaled~S Refaat, Nigamaa Nayakanti, Andre Cornman, Kan Chen, Bertrand Douillard, Chi~Pang Lam, Dragomir Anguelov, et~al.
\newblock Multipath++: Efficient information fusion and trajectory aggregation for behavior prediction.
\newblock In \emph{IEEE International Conference on Robotics and Automation (ICRA)}, 2022.

\bibitem[Vaswani et~al.(2017)Vaswani, Shazeer, Parmar, Uszkoreit, Jones, Gomez, Kaiser, and Polosukhin]{vaswani2017attention}
Ashish Vaswani, Noam Shazeer, Niki Parmar, Jakob Uszkoreit, Llion Jones, Aidan~N Gomez, {\L}ukasz Kaiser, and Illia Polosukhin.
\newblock Attention is all you need.
\newblock In \emph{Advances in Neural Information Processing Systems (NIPS)}, 2017.

\bibitem[Wang et~al.(2022)Wang, Ye, Gu, and Chen]{wang2022ltp}
Jingke Wang, Tengju Ye, Ziqing Gu, and Junbo Chen.
\newblock Ltp: Lane-based trajectory prediction for autonomous driving.
\newblock In \emph{Proceedings of the IEEE/CVF Conference on Computer Vision and Pattern Recognition (CVPR)}, 2022.

\bibitem[Wang et~al.(2023{\natexlab{a}})Wang, Zhu, Yu, Li, Ma, Jin, Ren, Ren, Wang, and Yang]{wang2023ganet}
Mingkun Wang, Xinge Zhu, Changqian Yu, Wei Li, Yuexin Ma, Ruochun Jin, Xiaoguang Ren, Dongchun Ren, Mingxu Wang, and Wenjing Yang.
\newblock Ganet: Goal area network for motion forecasting.
\newblock In \emph{IEEE International Conference on Robotics and Automation (ICRA)}, 2023{\natexlab{a}}.

\bibitem[Wang et~al.(2023{\natexlab{b}})Wang, Su, Da, and Yang]{wang2023prophnet}
Xishun Wang, Tong Su, Fang Da, and Xiaodong Yang.
\newblock Prophnet: Efficient agent-centric motion forecasting with anchor-informed proposals.
\newblock In \emph{Proceedings of the IEEE/CVF Conference on Computer Vision and Pattern Recognition (CVPR)}, 2023{\natexlab{b}}.

\bibitem[Ye et~al.(2021)Ye, Cao, and Chen]{ye2021tpcn}
Maosheng Ye, Tongyi Cao, and Qifeng Chen.
\newblock Tpcn: Temporal point cloud networks for motion forecasting.
\newblock In \emph{Proceedings of the IEEE/CVF Conference on Computer Vision and Pattern Recognition (CVPR)}, 2021.

\bibitem[Ye et~al.(2022)Ye, Xu, Xu, Wang, Cao, and Chen]{ye2022dcms}
Maosheng Ye, Jiamiao Xu, Xunnong Xu, Tengfei Wang, Tongyi Cao, and Qifeng Chen.
\newblock Dcms: Motion forecasting with dual consistency and multi-pseudo-target supervision.
\newblock \emph{arXiv preprint arXiv:2204.05859}, 2022.

\bibitem[Yuan et~al.(2021)Yuan, Weng, Ou, and Kitani]{yuan2021agentformer}
Ye Yuan, Xinshuo Weng, Yanglan Ou, and Kris~M Kitani.
\newblock Agentformer: Agent-aware transformers for socio-temporal multi-agent forecasting.
\newblock In \emph{Proceedings of the IEEE/CVF International Conference on Computer Vision (ICCV)}, 2021.

\bibitem[Zhan et~al.(2019)Zhan, Sun, Wang, Shi, Clausse, Naumann, Kummerle, Konigshof, Stiller, de~La~Fortelle, et~al.]{zhan2019interaction}
Wei Zhan, Liting Sun, Di Wang, Haojie Shi, Aubrey Clausse, Maximilian Naumann, Julius Kummerle, Hendrik Konigshof, Christoph Stiller, Arnaud de La~Fortelle, et~al.
\newblock Interaction dataset: An international, adversarial and cooperative motion dataset in interactive driving scenarios with semantic maps.
\newblock \emph{arXiv preprint arXiv:1910.03088}, 2019.

\bibitem[Zhang et~al.(2021)Zhang, Su, Hoang, Haynes, and Marchetti-Bowick]{zhang2021map}
Lingyao Zhang, Po-Hsun Su, Jerrick Hoang, Galen~Clark Haynes, and Micol Marchetti-Bowick.
\newblock Map-adaptive goal-based trajectory prediction.
\newblock In \emph{Conference on Robot Learning (CoRL)}, 2021.

\bibitem[Zhang et~al.(2024)Zhang, Liniger, Sakaridis, Yu, and Gool]{zhang2024real}
Zhejun Zhang, Alexander Liniger, Christos Sakaridis, Fisher Yu, and Luc~V Gool.
\newblock Real-time motion prediction via heterogeneous polyline transformer with relative pose encoding.
\newblock 2024.

\bibitem[Zhao et~al.(2021)Zhao, Gao, Lan, Sun, Sapp, Varadarajan, Shen, Shen, Chai, Schmid, et~al.]{zhao2021tnt}
Hang Zhao, Jiyang Gao, Tian Lan, Chen Sun, Ben Sapp, Balakrishnan Varadarajan, Yue Shen, Yi Shen, Yuning Chai, Cordelia Schmid, et~al.
\newblock Tnt: Target-driven trajectory prediction.
\newblock In \emph{Conference on Robot Learning (CoRL)}, 2021.

\bibitem[Zhou et~al.(2022)Zhou, Ye, Wang, Wu, and Lu]{zhou2022hivt}
Zikang Zhou, Luyao Ye, Jianping Wang, Kui Wu, and Kejie Lu.
\newblock Hivt: Hierarchical vector transformer for multi-agent motion prediction.
\newblock In \emph{Proceedings of the IEEE/CVF Conference on Computer Vision and Pattern Recognition (CVPR)}, 2022.

\bibitem[Zhou et~al.(2023)Zhou, Wang, Li, and Huang]{zhou2023query}
Zikang Zhou, Jianping Wang, Yung-Hui Li, and Yu-Kai Huang.
\newblock Query-centric trajectory prediction.
\newblock In \emph{Proceedings of the IEEE/CVF Conference on Computer Vision and Pattern Recognition (CVPR)}, 2023.

\bibitem[Zhu et~al.(2023)Zhu, Luan, and Shen]{zhu2023biff}
Yiyao Zhu, Di Luan, and Shaojie Shen.
\newblock Biff: Bi-level future fusion with polyline-based coordinate for interactive trajectory prediction.
\newblock In \emph{Proceedings of the IEEE/CVF Conference on Computer Vision and Pattern Recognition (CVPR)}, 2023.

\end{thebibliography}
